\documentclass[10pt,conference]{IEEEtran}
\usepackage{lipsum}

\usepackage{cite}

\ifCLASSINFOpdf
   \usepackage[pdftex]{graphicx}
   \graphicspath{{figs/}}
   \DeclareGraphicsExtensions{.pdf,.jpeg,.png}
\else
   \usepackage[dvips]{graphicx}
   \graphicspath{{../figs/}}
   \DeclareGraphicsExtensions{.eps}
\fi
\usepackage[cmex10]{amsmath}
\usepackage{amsthm}

\usepackage{algorithmic}

\usepackage{array}

\ifCLASSOPTIONcompsoc
  \usepackage[caption=false,font=normalsize,labelfont=sf,textfont=sf]{subfig}
\else
  \usepackage[caption=false,font=footnotesize]{subfig}
\fi

\usepackage{stfloats}
\usepackage{url}

\usepackage[normalem]{ulem}\usepackage{xcolor}

\newif\iffinal
\finalfalse

\iffinal
\else
\usepackage[switch]{lineno}
\fi

\begin{document}
%
\title{Combining YOLO and Visual Rhythm  for Vehicle Counting }



\author{
\IEEEauthorblockN{Victor Nascimento Ribeiro}
\IEEEauthorblockA{University of São Paulo - USP\\
Institute of Mathematics and Statistics\\
SP - São Paulo, Brazil}
\and
\IEEEauthorblockN{Nina S. T. Hirata}
\IEEEauthorblockA{University of São Paulo - USP\\
Institute of Mathematics and Statistics\\
SP - São Paulo, Brazil}
}


%


\maketitle

\begin{abstract}
Video-based vehicle detection and counting play a critical role in managing transport infrastructure. Traditional image-based counting methods usually involve two main steps: initial detection and subsequent tracking, which are applied to all video frames, leading to a significant increase in computational complexity. To address this issue, this work presents an alternative and more efficient method for vehicle detection and counting. The proposed approach eliminates the need for a tracking step and focuses solely on detecting vehicles in key video frames, thereby increasing its efficiency. To achieve this, we developed a system that combines YOLO, for vehicle detection, with Visual Rhythm, a way to create time-spatial images that allows us to focus on frames that contain useful information. Additionally, this method can be used for counting in any application involving unidirectional moving targets to be detected and identified. Experimental analysis using real videos shows that the proposed method achieves mean counting accuracy around 99.15\% over a set of videos, with a processing speed three times faster than tracking based approaches.
\end{abstract}


\IEEEpeerreviewmaketitle

\section{Introduction}
Due to continuous urban growth and population increase, there has been a significant rise in the number of vehicles circulating worldwide.
This, in turn, has led to the expansion of roads and highways to accommodate this growing traffic flow. However, the increase in the vehicle fleet makes it essential to implement an efficient and well-planned traffic control system to ensure the proper maintenance of an effective transportation infrastructure \cite{TrafficManagement}.

Systems based on computer vision and deep learning have become increasingly popular in the management of transportation infrastructure. This growing interest can be attributed to various factors, with emphasis on the wide availability of low-cost surveillance cameras, the ease of access to portable cameras and big data processing technologies \cite{Correlation}. Characteristics such as high precision and reduced costs make computer vision and deep learning an attractive and effective solution for addressing the challenges of traffic management in cities \cite{AplicationsDL}. In recent years, numerous studies have been conducted regarding efficient and accurate vehicle detection and counting.

In this paper, we present a more efficient approach for vehicle detection and counting in video captures. To accomplish this, we combine YOLO, a well known model for object detection, with Visual Rhythm, a technique that generates time-spatial images from videos. Unlike conventional methodologies that require vehicle detection and processing on a frame-by-frame approach, this integration enables us to selectively process frames containing relevant information, thereby enhancing both its efficiency and computational complexity at the cost of sacrificing real-time processing. 

\section{Background}
\label{sec-bg}
In this section, we provide essential background information for our proposed vehicle counting approach in video captures. We also contextualize our work within the relevant research landscape by reviewing prior studies in vehicle counting and Visual Rhythm.

\subsection{YOLO}

YOLO (You Only Look Once) is a real-time object detection 
deep learning model that has gained significant attention in computer vision. Starting with its 2015 release, YOLO has evolved to improve accuracy and performance, resulting in YOLOv8, launched in early 2023 \cite{yolo-v1 , yolo-evo}. Regarded as the current state-of-the-art for object detection, YOLO can be used to predict bounding boxes and class probabilities for multiple objects in an image. It also supports other tasks like object segmentation, pose estimation, tracking, and classification \cite{v8-github}.

This latest version has improved upon its predecessors by incorporating better feature extraction capabilities, allowing more accurate object detection in different environments \cite{v1-v8}. Additionally, it is an anchor-free model, so it can better predict bounding boxes for various object sizes and shapes, and also has a decoupled head to independently process classification and regression tasks.


\subsection{Visual Rhythm}

Let's consider a scenario where a static camera captures a video from a top-view perspective of vehicles. In this context, we can define the vehicle counting problem as the task of quantifying the number of vehicles that traverse a designated line (counting line) in the camera's field of view within a specific time interval. However, employing a frame-by-frame approach to solve this problem is not computationally efficient. In this approach, each frame in a video is processed independently, which leads to redundant computations and high memory usage.

Visual Rhythm (VR) generates time-spatial images from videos, a visual representation that merge spatial and temporal information from a video \cite{vr-first}. Designed for visual analysis, it can efficiently capture temporal information and summarize the video contents in a single image \cite{VR-definition}. This method enables to select frames featuring crucial visual content for counting and detection, thereby reducing the computational complexity tied to frame-by-frame processing of videos \cite{vr-video}.

\begin{figure}[htp]
    \centering
    \includegraphics[width=3.4in]{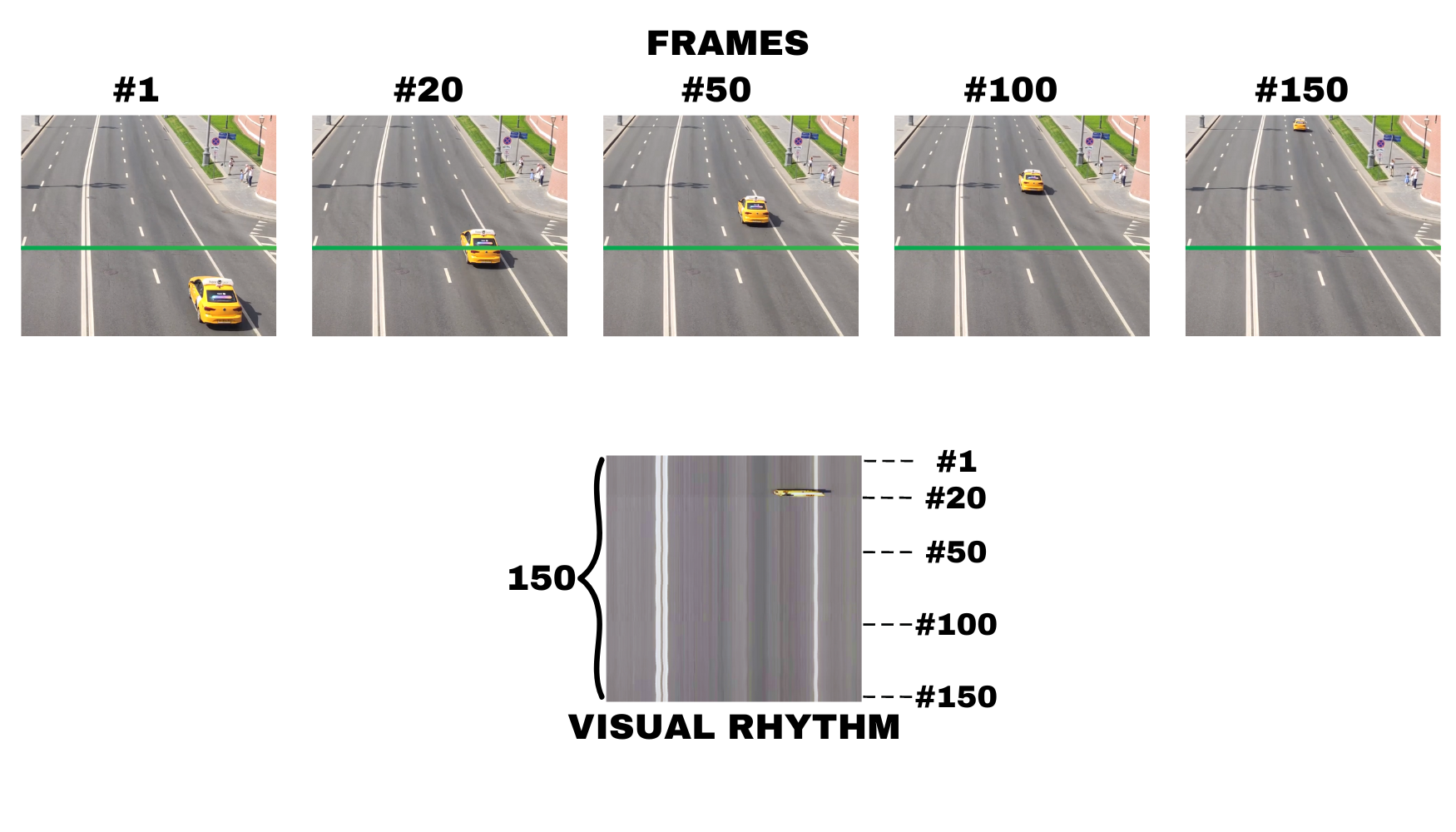}
    \caption{Visual Rhythm Generation.}
    \label{fig:vr-generation}
\end{figure}

Consider a video denoted as $f$ with $T$ frames of size $M \times N$.  The VR method is applied to each frame $f_1, \dots, f_T$, capturing exclusively the pixels along a predefined counting line. The collection of pixels from frame $t$, $t=1,\dots,T$, along this line is represented as ${vr}_t$. The VR image is formed by stacking all of them along the time axis to form an image of size $T \times N$, a time-spatial representation of the video, as show in Figure \ref{fig:vr-generation}. The top part shows five frames of a video sequence, where the "counting line" is superimposed in green (best viewed in electronic format), and the bottom part shows the time-spatial VR image (time dimension in the vertical axis) of the whole video sequence. Whenever a vehicle crosses the counting line, a mark is observed in the VR image, at the row corresponding to the frame index.

Applications using VR assume unidirectionally moving objects that cross the counting line in a top-view video and at a velocity within the camera's frame rate \cite{VR-plankton}.

\subsection{Related Works}
Currently, the majority of proposed methods for vehicle detection and counting using videos captured by static cameras are based on two steps: frame-by-frame detection and tracking. After detecting an object, tracking is initiated until it crosses the counting line, and then the count is updated in real-time. However, this algorithm incurs high computational costs as it detects and tracks each object individually within the frame. 

The work of Asha et al. \cite{Correlation} proposes a method for vehicle counting in a traffic video captured using portable cameras. YOLO is utilized for detection, while tracking employs a correlation filter with scale estimation. However, their method is restricted to counting vehicles within a single lane. Results indicate counting accuracy ranging from 92\% to 100\% across various videos.

In the work of Muhammad Azhad bin Zuraimi et al. \cite{DeepSORT}, the proposed system utilizes YOLO for vehicle detection and DeepSORT for vehicle tracking. The main difference of this work compared to the previously mentioned one is that it doesn't use a counting line, instead, vehicle counting occurs as they exit the frame or become no longer visible. The frame-by-frame approach leads this system to achieve more than 40 FPS when utilizing tiny versions of YOLO.

Researchers have used VR's capabilities to enhance visual analysis tasks by capturing temporal and spatial patterns in a single image. For instance, Matuszewski \cite{VR-plankton} utilized the VR approach for continuous plankton monitoring, showcasing its effectiveness in efficiently summarizing dynamic aquatic environments. Allan da Silva Pinto \emph{et al}. \cite{VR-definition} applied the method for video-based face spoofing detection. Torres \emph{et al}. \cite{vr-many} describe detectors derived from VR across three distinct computer vision tasks: abnormal event detection, human action classification, and gesture recognition.

\section{Proposed Method}
\begin{figure*}[!t]
    \centering
    \includegraphics[width=\textwidth]{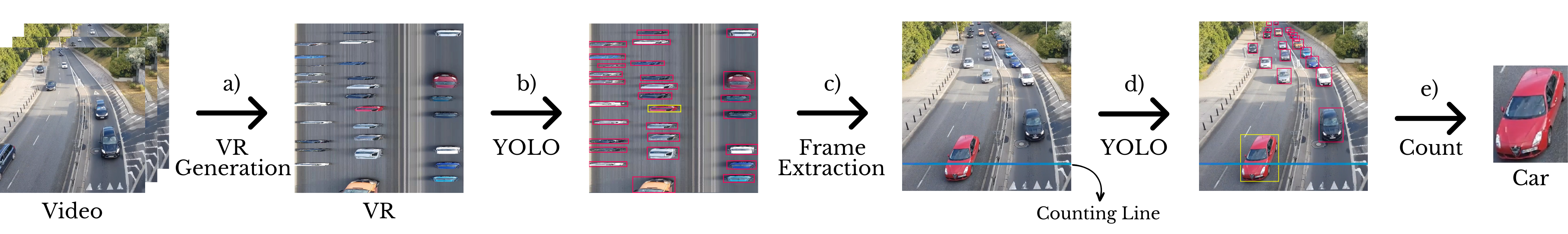}
    \caption{Data flow in the VR–based video counting vehicles.}
    \label{fig:method}
\end{figure*}

The proposed method combines YOLO and Visual Rhythm, allowing us to intelligently choose video frames with vital information for vehicle counting. 
Figure \ref{fig:method} illustrates the main steps of the proposed method.

First, in step (a), we create a Visual Rhythm image for segments of $T$ consecutive frames. Assuming a vehicle $i$ intersects the counting line in $T_i$ frames when crossing the counting line, the resultant VR image will contain a mark corresponding to vehicle $i$ with height equal to $T_i$ and width equal to the width of the vehicle. It is important to note, however, that certain marks in the VR image may not correspond to vehicles; for instance, any object crossing the counting line would also produce a corresponding mark. 

In the second step, we employ YOLO to detect each of the marks within the VR image (step (b)). Detecting these marks could be also achieved through background subtraction techniques. However, these methods could fail in handling varying weather conditions, sunlight intensity, and other sources of noise.

Next, for each mark in the VR image, we extract the corresponding frame from the video segment (step (c)). To that end, we know that the $y$ coordinate of the mark's center correspond to the temporal index of the frame when the vehicle's center crossed the counting line. In the diagram of Figure~\ref{fig:method} the extracted frame corresponds to the vehicle highlighted in yellow in the VR image.

After the relevant frame is obtained, we must certify that the mark in question indeed corresponds to a vehicle. Thus, in step (d) we use YOLO to detect all vehicles in the extracted frame and then, for each detected vehicle, we compute the distance between the $x$ coordinates of the vehicle's bounding box and the $x$ coordinates of the mark in the VR image. The vehicle that best matches the size and position of the mark is selected as the one corresponding to the mark. If there are more than one, the one with closest $y$ coordinate is selected as the vehicle corresponding to the mark, and the counting of the vehicle class is updated (step (e)).'

\begin{figure}[htp]
    \centering
    \includegraphics[width=3.25in]{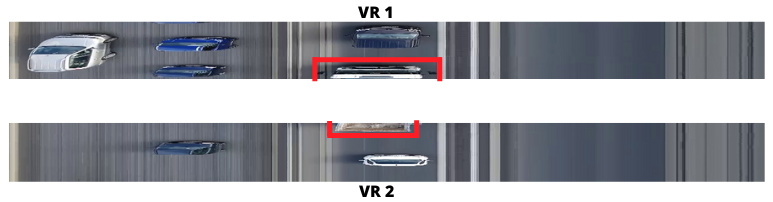}
    \caption{Vehicle represented by distinct marks in two consecutive VR images.}
    \label{fig:double-count}
\end{figure}

It is important to note that the above method is applied on sequences of length $T$ in order to keep the VR image size manageable by YOLO. Thus, to process long video sequences, they are partitioned into contiguous and non-overlapping segments of length $T$.
This approach can sometimes lead to situations where the same vehicle appears in the VR images of two consecutive segments. 
To avoid double counting, we implement a verification process. Specifically, we check if the coordinates of a mark fall along the lower edge of a VR image; if they do, we store these coordinates. As we proceed to create and analyze the subsequent VR, we examine whether any mark appear along the upper edge of the image. 
If so, and if its $x$ coordinate match the ones stored previously, then we disregard this mark, as it corresponds to the same vehicle in the preceding VR image. This verification process is carried out for each generated VR image. Figure \ref{fig:double-count} shows a visual representation of this situation, where we see that the center coordinates of an mark is within the other mark coordinates. 

\section{Results and discussion}
To implement the proposed vehicle detection and counting method, we used a computer equipped with a GeForce GTX 1080 Ti running on an Ubuntu operating system. The implementation was carried out using Python, utilizing the Ultralytics and OpenCV libraries.

In this study, we ultilized YOLOv8-small. To enhance its vehicle detection performance, we employed transfer learning by fine-tuning its pre-trained model in a task-specific dataset. This technique was applied to detecting marks and vehicles.

\subsection{Datasets and training}

We use 4 publicly available videos captured from a top-view perspective by a static camera, in diverse locations and scenarios. This highlights the variation in camera angles, the number of lanes for vehicles, and their directions.
To fine-tune YOLO for vehicle detection, we built a dataset consisting of 960 frames randomly selected from these videos, as outlined in Table \ref{tab:dataset}.
We defined 6 vehicle classes: Bus, Car, Motorbike, Pickup, Truck, and Van, and annotated all occurrences of instances of these classes in these selected frames 
using the Roboflow's framework \cite{roboflow}.

\begin{table}[htp]
\renewcommand{\arraystretch}{1.3}
    \caption{Description of the Dataset's Image Distribution}
    \label{tab:dataset}
    \centering
    \begin{tabular}{|c||c||c|}
        \hline
        Video & Video Length (frames) & \# of Selected Frames\\
        \hline
        1 & 720 & 60\\ 
        \hline
        2 & 9180 & 300\\ 
        \hline
        3 & 20340 & 300\\ 
        \hline
        4 & 62880 & 300\\ 
    \hline
    \end{tabular}
\end{table}

\begin{figure}[htp]
    \centering
    \includegraphics[width=2.5in]{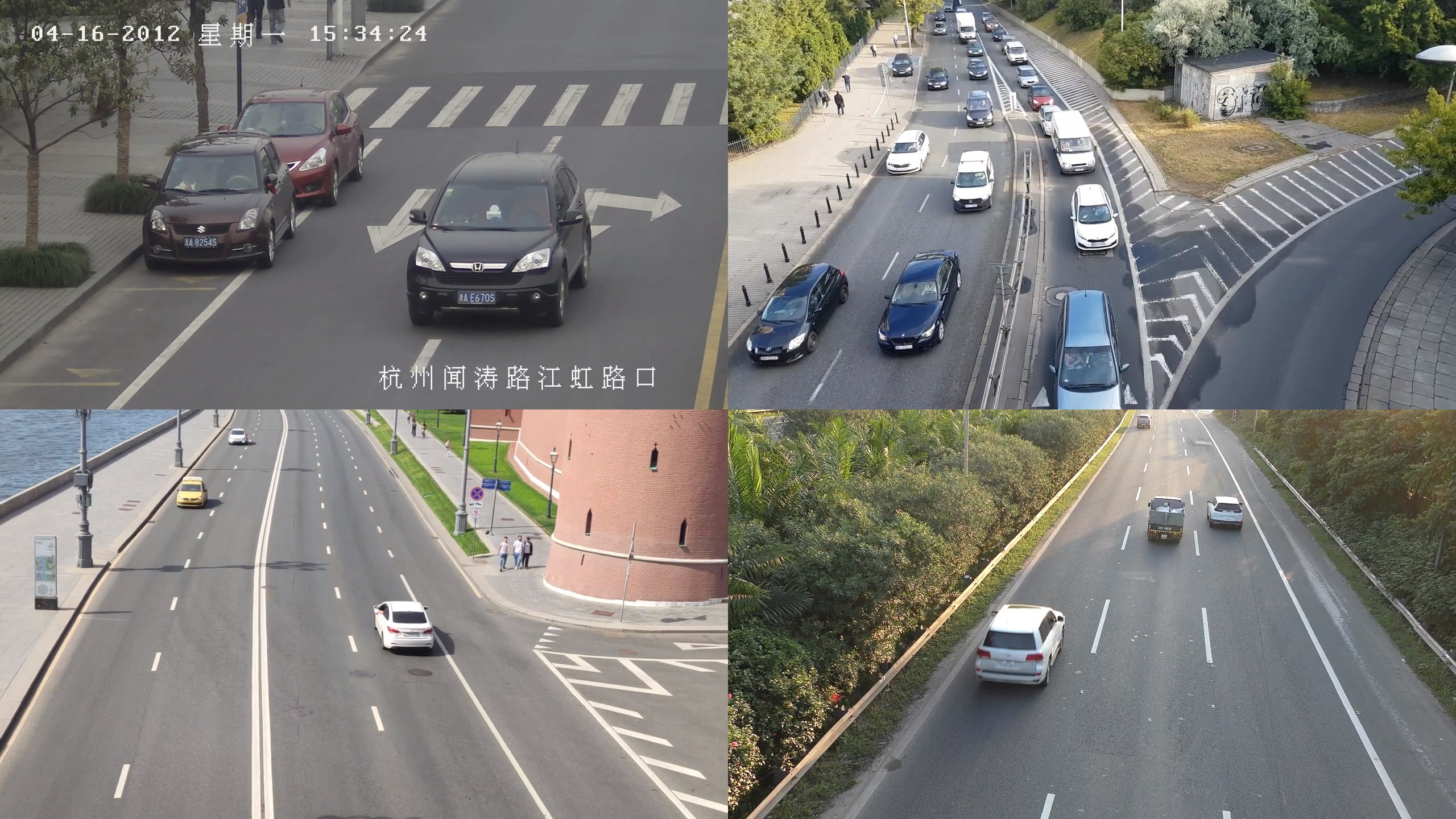}
    \caption{Images samples from the dataset}
    \label{fig:dataset}
\end{figure}

All frame images were resized to a uniform 1280$\times$720 resolution, without any additional preprocessing or augmentation. Refer to Figure \ref{fig:dataset} for frame samples of the dataset.

The dataset was partitioned into training, validation, and testing subsets. To that end, from each video, the first 70\% of the selected frames were included in the training set, the next 20\% in the validation set, and the last 10\% to the testing set. We also took care to ensure that a same vehicle is not included in more than one of the sets. 

To train the model for mark detection, we generated 33 VR images corresponding to segments of $T=900$ frames (30 seconds) around the frames in the training and validation sets for vehicle detection described above. The marks have been manually annotated in these VR images. We further augmented our dataset by applying horizontal and vertical flips, along with cropping, resulting in a total of 79 images.

For training, we set up 60 epochs with a batch size of 8 for mark detection, and 100 epochs with a batch size of 64 for vehicle detection, using YOLOv8's standard configuration for other hyperparameters. The mark detection model achieved a final Mean Average Precision (mAP) of 0.99136 at IoU 0.5 and 0.60865 at IoU 0.5:0.95 on the validation set. For vehicle detection, the results were mAP of 0.84274 at IoU 0.5 and 0.69166 at IoU 0.5:0.95, also on the validation set.

\subsection{Results}
In our experiments, we used segments with length $T=900$ frames (30 seconds) to build the VR images, an optimal size for YOLO processing. The counting line was positioned at a height of 120 pixels on the original videos, ensuring coverage of the entire vehicle path. We compare our approach efficiency to Roboflow's frame-by-frame vehicle counting and detection system \cite{roboflow}. This system employs ByteTrack for tracking and supervision for real-time object counting. For both systems, identical environments and model weights were utilized. 
Table \ref{tab:test1} shows a comparison of the two methods  on the test segment across various videos, with counting accuracy computed regardless of vehicle class. We did not include the first video due to its short duration.

\begin{table}[htp]
\renewcommand{\arraystretch}{1.3}
    \caption{Counting Accuracy in Each Video}
    \label{tab:test1}
    \centering
    \begin{tabular}{|c||c||c||c||c|}
        \hline
        System & Frame rate & Video 2 & Video 3 & Video 4\\
        \hline
        Visual Rhythm & 186 FPS & 100\% & 98.90\%  & 98.56\% \\
        \hline
        ByteTrack & 56 FPS & 100\% & 99.04\% & 98.48\%\\
    \hline
    \end{tabular}
\end{table}

The results demonstrate that our Visual Rhythm-based method presented an efficiency improvement of approximately three times when compared to the traditional detection and tracking method. This efficiency gain is largely attributed to the elimination of the need for frame-by-frame processing, which significantly contributes to enhancing overall system performance. Moreover, it's worth noting that the generation of VR images, a crucial aspect of our approach, is not overly computationally demanding. More importantly, the gain in efficiency did not impact counting accuracy. We also analyze the results of these approaches when determining the vehicle classes while performing counting. Table \ref{tab:test2} shows our approach's classification accuracy over the test set (four videos). 

\begin{table}[htp]
\renewcommand{\arraystretch}{1.3}
    \caption{Classification Accuracy in Each Video on VR Approach}
    \label{tab:test2}
    \centering
    \begin{tabular}{|c||c||c||c||c||c|}
        \hline
        Car & Bus & Motorbike & Pickup & Truck & Van\\
        \hline
        99.1\% & 86.0\% & 81.6\% & 77.0\% & 96.2\% & 92.9\% \\
    \hline
    \end{tabular}
\end{table}

Remarkably, the "Car" category exhibits the highest accuracy, with an impressive 99.1\% accuracy rate. However, classifying "Pickup" and "Motorbike" proves comparatively more challenging, with accuracies of 77.0\% and 81.6\%, respectively, which are low. 
We note that the accuracies were computed taking into consideration only the detected and counted vehicles. Classes that exhibit lower accuracies are those less frequent in the dataset.

\section{Conclusion}
The initial hypothesis pointing that the fusion of YOLO with Visual Rhythm for vehicle counting and detection would enhance system performance has been confirmed. Our approach outperforms traditional methods, with a speed-up of approximately 3 times. This show the efficacy of discarding the frame-by-frame processing. Moreover, our approach maintains approximately the same accuracy levels as tracking methods, effectively capturing vehicles through Visual Rhythm but has the disadvantage of not being a real-time application. 

Applying transfer learning to fine-tune YOLO's pre-trained model for task of vehicle detection show satisfactory results in terms of mAP . However, as this metric is calculated across all possible classes, the model's limitations become apparent when analyzing the details presented in Table \ref{tab:test2}. Therefore, there is room for further improvement in the model's performance, particularly by including more examples from classes with low classification accuracy.


\section*{Acknowledgment}

The authors thank support from MCTI (\textit{Ministério da Ciência, Tecnologia e Inovações, Brazil}), law 8.248, PPI-Softex - TIC 13 - 01245.010222/2022-44, \textit{Fundação de Apoio à Universidade de São Paulo} (FUSP), and FAPESP (grant 2015/22308-2).

\bibliographystyle{IEEEtran}
\bibliography{example}

\begin{thebibliography}{10}
\providecommand{\url}[1]{#1}
\csname url@samestyle\endcsname
\providecommand{\newblock}{\relax}
\providecommand{\bibinfo}[2]{#2}
\providecommand{\BIBentrySTDinterwordspacing}{\spaceskip=0pt\relax}
\providecommand{\BIBentryALTinterwordstretchfactor}{4}
\providecommand{\BIBentryALTinterwordspacing}{\spaceskip=\fontdimen2\font plus
\BIBentryALTinterwordstretchfactor\fontdimen3\font minus
  \fontdimen4\font\relax}
\providecommand{\BIBforeignlanguage}[2]{{%
\expandafter\ifx\csname l@#1\endcsname\relax
\typeout{** WARNING: IEEEtran.bst: No hyphenation pattern has been}%
\typeout{** loaded for the language `#1'. Using the pattern for}%
\typeout{** the default language instead.}%
\else
\language=\csname l@#1\endcsname
\fi
#2}}
\providecommand{\BIBdecl}{\relax}
\BIBdecl

\bibitem{TrafficManagement}
A.~A. Kurzhanskiy and P.~Varaiya, ``Traffic management: An outlook,''
  \emph{Economics of Transportation}, vol.~4, no.~3, pp. 135--146, 2015.

\bibitem{Correlation}
C.~S. Asha and A.~V. Narasimhadhan, ``Vehicle counting for traffic management
  system using yolo and correlation filter,'' in \emph{IEEE International
  Conference on Electronics, Computing and Communication Technologies
  (CONECCT)}, 2018, pp. 1--6.

\bibitem{AplicationsDL}
P.~Patil, ``Applications of deep learning in traffic management: A review,''
  \emph{International Journal of Business Intelligence and Big Data Analytics},
  vol.~5, no.~1, p. 16–23, Jan. 2022.

\bibitem{yolo-v1}
J.~Redmon, S.~Divvala, R.~Girshick, and A.~Farhadi, ``You only look once:
  Unified, real-time object detection,'' 2016.

\bibitem{yolo-evo}
P.~Jiang, D.~Ergu, F.~Liu, Y.~Cai, and B.~Ma, ``A review of yolo algorithm
  developments,'' \emph{Procedia Computer Science}, vol. 199, pp. 1066--1073,
  2022, the 8th International Conference on Information Technology and
  Quantitative Management (ITQM 2020 and 2021): Developing Global Digital
  Economy after COVID-19.

\bibitem{v8-github}
\BIBentryALTinterwordspacing
G.~Jocher, A.~Chaurasia, and J.~Qiu, ``{YOLO by Ultralytics},'' Jan. 2023.
  [Online]. Available: \url{https://github.com/ultralytics/ultralytics}
\BIBentrySTDinterwordspacing

\bibitem{v1-v8}
J.~Terven and D.~Cordova-Esparza, ``A comprehensive review of yolo: From yolov1
  and beyond,'' 2023.

\bibitem{vr-first}
S.~Guimar, M.~Couprie, N.~Leite, and D.~A. Araujo, ``A method for cut detection
  based on visual rhythm,'' in \emph{Proceedings XIV Brazilian Symposium on
  Computer Graphics and Image Processing}, 2001, pp. 297--304.

\bibitem{VR-definition}
A.~d.~S. Pinto, H.~Pedrini, W.~Schwartz, and A.~Rocha, ``Video-based face
  spoofing detection through visual rhythm analysis,'' in \emph{25th SIBGRAPI
  Conference on Graphics, Patterns and Images}, 2012, pp. 221--228.

\bibitem{vr-video}
K.-d. Seo, S.~J. Park, and S.-h. Jung, ``Wipe scene-change detector based on
  visual rhythm spectrum,'' \emph{IEEE Transactions on Consumer Electronics},
  vol.~55, no.~2, pp. 831--838, 2009.

\bibitem{VR-plankton}
D.~J. Matuszewski, ``Computer vision for continuous plankton monitoring,''
  Master's thesis, Instituto de Matemática e Estatística, University of São
  Paulo, São Paulo, 2014, retrieved 2023-08-06.

\bibitem{DeepSORT}
M.~A. Bin~Zuraimi and F.~H. Kamaru~Zaman, ``Vehicle detection and tracking
  using yolo and deepsort,'' in \emph{IEEE 11th IEEE Symposium on Computer
  Applications and Industrial Electronics (ISCAIE)}, 2021, pp. 23--29.

\bibitem{vr-many}
\BIBentryALTinterwordspacing
B.~S. Torres and H.~Pedrini, ``Detection of complex video events through visual
  rhythm,'' \emph{The Visual Computer}, vol.~34, no.~2, pp. 145--165, 02 2018.
  [Online]. Available: \url{https://doi.org/10.1007/s00371-016-1321-1}
\BIBentrySTDinterwordspacing

\bibitem{roboflow}
B.~Dwyer, J.~Nelson, J.~Solawetz \emph{et~al.}, ``Roboflow (version 1.0)
  [software],'' \url{https://roboflow.com}, 2022, computer vision.

\end{thebibliography}
%
%


\end{document}